\documentclass[letterpaper, 10pt, conference]{ieeeconf}

\usepackage{amsmath, amssymb, amsfonts, mathtools}
\usepackage{bm}
\usepackage{array, float, xcolor}
\usepackage{graphicx}
\usepackage{dsfont}
\usepackage[T1]{fontenc}
\usepackage{makecell}
\usepackage{booktabs}
\allowdisplaybreaks

\IEEEoverridecommandlockouts

\usepackage{amsthm}
\DeclareMathOperator*{\argmin}{arg\,min}
\let\labelindent\relax
\usepackage{enumitem}

\newtheorem{theorem}{Theorem}
\newtheorem{lemma}[theorem]{Lemma}
\newtheorem{proposition}[theorem]{Proposition}

\theoremstyle{definition}

\newtheorem{assumption}{Assumption}
\newtheorem{problem}{Problem}
\theoremstyle{remark}

\makeatletter
\newcommand{\objitem}[2]{\item[\textbf{#1}]\edef\@currentlabel{#1}\label{#2}}
\makeatother

\usepackage{cite}
\makeatletter
\let\NAT@parse\undefined
\makeatother
\usepackage[colorlinks=true, linkcolor=blue, citecolor=blue, urlcolor=blue, pdfborder={0 0 0}]{hyperref}
\usepackage[nameinlink,capitalise]{cleveref}
\crefname{assumption}{Assumption}{Assumptions}
\Crefname{assumption}{Assumption}{Assumptions}
\crefname{problem}{Problem}{Problems}
\Crefname{problem}{Problem}{Problems}
\crefname{definition}{Definition}{Definitions}
\crefname{remark}{Remark}{Remarks}
\crefname{proposition}{Proposition}{Propositions}
\crefname{lemma}{Lemma}{Lemmas}
\hypersetup{
    pdfauthor={Hamza Tariq and Adeel Akhtar},
    pdftitle={Dual Path Following for an Aerial Manipulator Under State and Input Constraints via Transverse Feedback Linearization},
    pdfsubject={Safety-Critical Control, Path Following, Aerial Manipulation},
    pdfkeywords={aerial manipulation, path following, transverse feedback linearization, control barrier functions}}
\usepackage{macros}
\usepackage{url}
\newcommand{\pc}{p_{c}}
\newcommand{\pe}{p_{e}}
\newcommand{\ud}{u_{d}}
\newcommand{\ut}{u_{t}}
\newcommand{\utmin}{\underline{u}_{t}}
\newcommand{\utmax}{\overline{u}_{t}}
\newcommand{\detJ}{\det J_{\rho}}
\newcommand{\rdel}{r_{\delta}}

\begin{document}

\title{Dual Path Following for an Aerial Manipulator Under State and Input Constraints via Transverse Feedback Linearization}

\author{Hamza Tariq$^{1}$ and Adeel Akhtar$^{1}$%
\thanks{$^{1}$Hamza Tariq and Adeel Akhtar are with the Department of Mechanical \& Industrial Engineering at the New Jersey Institute of Technology, Newark, NJ 07102, USA.
        {\tt\small \{ht299, adeel.akhtar\}@njit.edu}}%
}

\maketitle

\begin{abstract}
\looseness=-1 This paper studies dual path following for a quadrotor carrying a three-degree-of-freedom arm. The center of mass of the combined system must follow a vehicle path and the end effector must follow a second path, while the arm respects its joint limits and avoids its singular postures. The vehicle and the arm are mechanically coupled, so one controller must handle both paths. With the center of mass and the end-effector path coordinates as outputs, and a dynamic extension of the thrust, the system admits a normal form without internal dynamics. We characterize the states where this normal form fails. We formulate the transverse feedback linearizing controller as a quadratic program that relaxes only the speed of the end effector along its path and imposes safety and actuator limits as inequality constraints. The program reduces to a scalar problem with a closed-form solution. If it remains strictly feasible for all time, the feedback is locally Lipschitz, both paths are invariant and attractive, and every constraint holds. In the physics engine Drake, the controller resolves a conflict between the two speed references by slowing the end effector while both paths remain invariant. By contrast, a cascade with a safety filter leaves both paths. Code and animations are available at \url{https://gradslab.github.io/am_dual_pf-/}.
\end{abstract}

\section{Introduction}
\looseness=-1 Aerial manipulators, that is, multirotors carrying a robotic arm, are used for the inspection and maintenance of structures such as bridges, pipes and tanks~\cite{ollero2021past}. These tasks may require the end effector to trace a prescribed route along the structure while the vehicle follows a second route that keeps its rotors clear of the structure. The end effector and the vehicle must stay on their routes. The speed along them is secondary, and it may be reduced when the arm approaches a joint limit or a singular posture~\cite[Sec.~3.3]{siciliano2009robotics}, a configuration at which the joints can no longer move the end effector in every direction relative to the center of mass.

\looseness=-1 In this work we study the \emph{dual path-following} problem. Its objective is convergence of the vehicle and the end effector to their paths together with \emph{path invariance}, that is, once they reach their paths they remain on them for all future time. Transverse feedback linearization (TFL)~\cite{nielsen2008local} achieves this by rendering the dynamics transverse to the path linear and controllable, so that a linear feedback makes the path invariant and attractive~\cite{akhtar2015path}. Path invariance via TFL has been shown for a single quadrotor~\cite{akhtar2012path, allawati2026qstfl}. A partial feedback linearization gives it for a manipulator on a wheeled mobile base~\cite{gill2017path}. There, only the end effector is assigned a path, and the dynamics of the base and of the arm are assumed to be decoupled. In \cite{li2016synchronized}, a differential drive robot and a serial manipulator, two separate systems, are synchronized while each follows its own path. An aerial manipulator allows neither simplification. The vehicle and the arm form one mechanically coupled system. The arm moves the center of mass and the inertia of the vehicle, while the attitude moves the end effector. One system must therefore follow both paths.

\begin{figure}[!t]
\centering
\includegraphics[width=0.85\linewidth]{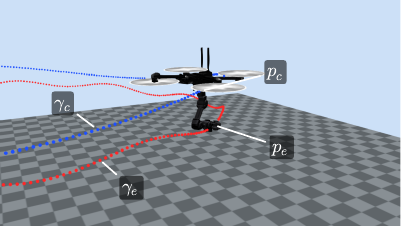}

\vspace{-0.1in}

\caption{Dual path following in Drake. The center of mass $\pc$ of the combined system follows the vehicle path $\gamma_c$ (blue, dotted) and the end effector $\pe$ follows the end-effector path $\gamma_e$ (red, dotted). The solid red line is the recent end-effector trajectory.}

\vspace{-0.08in}

\label{fig:concept}
\end{figure}

\looseness=-1 Control of aerial manipulators is surveyed in~\cite{ollero2021past}. Differential-flatness-based controllers and planners take the center of mass, or its momentum, with the yaw and the joint angles as flat outputs \cite{welde2021dynamically, wei2021differential}. These outputs have a well-defined vector relative degree after a dynamic extension of the thrust \cite{wei2021differential}. The end-effector pose is flat only for designs whose center of mass is fixed in the end-effector frame \cite{welde2020coordinate}. Kinematic controllers make the end effector of a hexarotor with a three-degree-of-freedom (3-DOF) arm follow a path \cite{carvajal2023path, carvajal2024multitask}, but without path invariance or safety constraints. A feedback linearizing controller tracks trajectories of the base and the end effector together~\cite{mirnajafizadeh2025unified}, again without constraints. A quadratic program (QP) with actuator bounds and joint limits tracks the base position, the yaw and the end effector simultaneously \cite{laribi2026qp}, but without asymptotic tracking guarantees. To the best of our knowledge, no existing controller for an aerial manipulator renders both a vehicle path and an end-effector path invariant under safety constraints.

\looseness=-1 In~\cite{tariq2026relaxed} we formulate the feedback linearizing controller of a unicycle as a QP with relaxed equality constraints. In~\cite{tariq2026decentralized} we present the TFL controller of a quadrotor as a QP and relax only its along-path speed constraint, just as the stability constraint is relaxed in~\cite{ames2016control}. This preserves the path and heading dynamics exactly, while control barrier function (CBF) inequalities enforce safety. In this work we extend the relaxed QP-based TFL to a quadrotor carrying a 3-DOF arm~(\cref{fig:concept}). The arm moves the center of mass and brings joint limits and singular postures that the quadrotor alone does not have. The main contributions are the following.
\begin{enumerate}[leftmargin=*,nosep]
\item We propose an output, built on the flat outputs of \cite{welde2021dynamically, wei2021differential}, for which the dynamically extended aerial manipulator model admits a normal form without internal dynamics, and we give its singular set in closed form (\cref{prop:rd,prop:arm}).
\item We propose the TFL controller as a QP that imposes safety and actuator limits and relaxes only the speed of the end effector along its path. We show that this QP has a closed-form, locally Lipschitz solution (\cref{thm:lipschitz}).
\item We prove that, under strict feasibility, both paths are invariant and attractive and every constraint holds. On closed paths at low vehicle speed, a feasibility margin along the solution extends these guarantees to all time (\cref{thm:safety}).
\end{enumerate}
We evaluate the controller against the nominal controller and a safety-filtered cascade in Drake \cite{drake}, on a Skydio X2 with an OpenMANIPULATOR-X arm.

\textit{Notation:} A tuple $(x_1,\dots,x_k)$ of vectors is the column vector obtained by stacking them, and a tuple of matrices with the same number of columns is their vertical stack. We write $\norm{\cdot}$ for the Euclidean norm, $M\succ0$ means that $M$ is symmetric positive definite, $\mathrm{blkdiag}(\cdot)$ is the block-diagonal matrix of its arguments, and $e_i$ is the $i$th standard basis vector. For $a\in\Real^3$, $\hat{a}$ is the matrix with $\hat{a}b = a\times b$. For a map $\varphi$, $\partial\varphi/\partial x$ is its Jacobian. For a function $h$ and vector fields $F$ and $G$, $L_Fh$ is the Lie derivative, $L_F^0h := h$, $L_F^kh := L_F(L_F^{k-1}h)$ and $L_GL_F^kh := L_G(L_F^kh)$.

\section{Preliminaries}
\subsection{Exponential Control Barrier Functions~(ECBF)}\label{sec:ecbf}
\looseness=-1 Consider a control-affine system $\dot{x} = F(x) + G(x)u$ with state $x\in\Real^n$, input $u\in\Real^p$ and locally Lipschitz $F$ and $G$. Let $h:\Real^n\to\Real$ be a smooth function with $L_Gh = 0$. Let $k_1,k_2>0$ be such that the polynomial $s^2 + k_2 s + k_1$ has negative real roots $\nu_1,\nu_2$. The ECBF condition of \cite{nguyen2016exponential} is
\begin{equation}\label{eq:ecbf}
L_F^2 h + L_GL_Fh\,u \ge -k_1h - k_2L_Fh,
\end{equation}
and the admissible set of $h$ is $\mathcal{A} := \{x\in\Real^n : h(x)\ge0,\ L_Fh(x) - \nu_1 h(x)\ge 0\}$. If a solution satisfies \eqref{eq:ecbf} on $[0,t_f)$ and $x(0)\in\mathcal{A}$, then $x(t)\in\mathcal{A}$ for all $t\in[0,t_f)$ \cite[Thm.~1]{nguyen2016exponential}.

\subsection{Aerial Manipulator Model}
\looseness=-1 We consider a quadrotor carrying a 3-DOF arm with revolute joints on its underside. Let $\mathcal{W}$ be the inertial frame, with $e_3$ pointing up, and $\mathcal{B}$ the body frame of the quadrotor, with origin $p_b\in\Real^3$ on the thrust axis and third axis along it. The attitude $R\in\SO{3}$ of $\mathcal{B}$ relative to $\mathcal{W}$ is parameterized by the roll, pitch and yaw angles $\Phi := (\phi,\theta,\psi)$ as $R = R_z(\psi)R_y(\theta)R_x(\phi)$. The angular velocity $\omega\in\Real^3$ is expressed in $\mathcal{B}$, so that $\dot R = R\hat\omega$ and $\dot\Phi = Y(\Phi)\omega$. The matrix $Y(\Phi)$ is defined and invertible for $\cos\theta\neq0$ \cite[Sec.~3.6]{siciliano2009robotics}. 

\looseness=-1 The arm has a waist, a shoulder and an elbow joint, the usual configuration in aerial manipulation \cite{ollero2021past}. The waist axis is parallel to the thrust axis. The shoulder and elbow axes are parallel to each other and perpendicular to the waist axis, so the upper arm and the forearm move in a plane that contains the waist axis. The joint angles of the waist, shoulder and elbow are $\vartheta := (\vartheta_1,\vartheta_2,\vartheta_3)\in\Real^3$ and the configuration is $q := (p_b,\Phi,\vartheta)\in\Real^9$. The inputs are the collective thrust $\ut\in\Real$ along the thrust axis, the body moments $\tau\in\Real^3$ and the joint torques $\tau_\vartheta\in\Real^3$. Thus, the system has seven inputs and nine degrees of freedom, making it underactuated.

\looseness=-1 Let $m$ be the total mass of the combined system. The forward kinematics of the arm give smooth maps $r_c:\Real^3\to\Real^3$ and $r_e:\Real^3\to\Real^3$ from the joint angles $\vartheta$ to the positions in $\mathcal{B}$ of the center of mass of the combined system and of the end effector. In $\mathcal{W}$ these positions are $\pc := p_b + R\,r_c(\vartheta)$ and $\pe := p_b + R\,r_e(\vartheta)$. We neglect aerodynamic forces, so thrust and gravity are the only external forces. Let $\rho := r_e - r_c$ be the position of the end effector relative to the center of mass in $\mathcal{B}$, with Jacobian $J_\rho(\vartheta) := \partial\rho/\partial\vartheta\in\Real^{3\times3}$. The equations of motion split into Newton's law for the center of mass and the equations of the rotational and joint motion about it, as in \cite{welde2021dynamically}:
\begin{subequations}\label{eq:model}
\begin{align}
m\ddot{\pc} &= \ut R e_3 - m g e_3, \label{eq:com}\\
M_r(\vartheta)\begin{bmatrix}\dot\omega\\ \ddot\vartheta\end{bmatrix} &= \begin{bmatrix}\tau\\ \tau_\vartheta\end{bmatrix} + \ut\Upsilon(\vartheta) - c(\vartheta,\omega,\dot\vartheta), \label{eq:rot}
\end{align}
\end{subequations}
\looseness=-1 Here $M_r(\vartheta)\succ0$ is the inertia matrix of the rotational and joint motion about the center of mass, $c$ collects the Coriolis and centrifugal terms, and $\Upsilon(\vartheta) := \big(e_3\times r_c(\vartheta),\ -(\partial r_c/\partial\vartheta)^\top e_3\big)$ is the generalized force of a unit thrust, which acts at $p_b$ and not at $\pc$. Gravity acts at the center of mass, so it does not appear in \eqref{eq:rot}. Equations \eqref{eq:model} are equivalent to the Euler--Lagrange equations in the coordinates $q$,
\begin{equation}\label{eq:el}
M(q)\ddot q + C(q,\dot q)\dot q + g(q) = B_t(q)\,\ut + B_\tau(q)\,(\tau,\tau_\vartheta),
\end{equation}
\looseness=-1 with inertia matrix $M(q)$, positive definite for $\cos\theta\neq0$, Coriolis matrix $C(q,\dot q)$, gravity vector $g(q)$, $B_t(q) := (Re_3, 0, 0)\in\Real^9$ and $B_\tau(q) := \begin{bmatrix} 0_{6\times3} & \mathrm{blkdiag}(Y(\Phi)^{-1}, I_3)\end{bmatrix}^\top\in\Real^{9\times6}$. 

As in \cite{akhtar2012path}, we add two integrators to the thrust as a dynamic extension, so that the virtual output of \cref{sec:normal} has a well-defined vector relative degree. Define $(x_1,\dots,x_{18}) := (q,\dot q)$, $x_{19} := \ut$ and $x_{20} := \dot\ut$, so that the new thrust input is $\ud := \ddot\ut$. The extended state is $x := (x_1,\dots,x_{20})\in\Real^{20}$ and the extended input is $u := (\ud,\tau,\tau_\vartheta)\in\Real^7$. Solving \eqref{eq:el} for $\ddot q$ gives the extended model
\begin{equation}\label{eq:affine}
\dot x = F(x) + G(x)u,
\end{equation}
with $F(x) = \big(\dot q,\ M^{-1}(B_t x_{19} - C\dot q - g),\ x_{20},\ 0\big)$ and $G(x) = \begin{bmatrix} e_{20} & \bar G\end{bmatrix}$, where $\bar G := \begin{bmatrix} 0_{6\times9} & B_\tau^\top M^{-1} & 0_{6\times2}\end{bmatrix}^\top$ and the arguments $q$ and $\dot q$ are omitted. The maps $F$ and $G$ are smooth on the set where $\cos\theta\neq0$.

\section{Problem Formulation}\label{sec:problem}
\looseness=-1 Let $\gamma_c\subset\Real^3$ be the vehicle path and $\gamma_e\subset\Real^3$ the end-effector path. As in \cite{akhtar2012path}, each $\gamma_j$, $j\in\{c,e\}$, is a smooth embedded curve with a regular parameterization $\sigma_j:D_j\to\Real^3$ on an open interval $D_j\subset\Real$. If $\gamma_j$ is closed, $D_j = \Real$ and $\sigma_j$ is $L_j$-periodic. Each path is also the zero level set $\gamma_j = \{p : s_j(p) = 0\}$ of a smooth map $s_j:\Real^3\to\Real^2$ with $\rank\,\partial s_j/\partial p = 2$ on $\gamma_j$. Set $\mathcal{C}_j := D_j$ if $\gamma_j$ is not closed and $\mathcal{C}_j := \Real/L_j\mathbb{Z}$ if it is. Let $U_{\gamma_j}\subset\Real^3$ be a tubular neighborhood of $\gamma_j$ \cite[Ch.~2]{guillemin1974differential} and $\varpi_j:U_{\gamma_j}\to\mathcal{C}_j$ a smooth map with $\varpi_j(\sigma_j(\ell)) = \ell$ for every $\ell\in D_j$, modulo $L_j$ for closed paths. For closed paths, the derivatives of $\varpi_j$ are computed with local real-valued lifts, whose derivatives agree on overlaps.

The Jacobian $\Pi_j := \partial(s_j,\varpi_j)/\partial p\in\Real^{3\times3}$ is nonsingular on $\gamma_j$. Indeed, the two rows of $\partial s_j/\partial p$ are independent and annihilate the tangent $\sigma_j'$, since $s_j$ vanishes along $\gamma_j$, while $(\partial\varpi_j/\partial p)\,\sigma_j' = 1$. We shrink $U_{\gamma_j}$ to $\{p\in U_{\gamma_j} : \norm{s_j(p)} < \delta_j\}$ with $\delta_j>0$ such that $\Pi_j$ is nonsingular on it, which we assume is possible when $\gamma_j$ is not closed. When $\gamma_j$ is closed, it is compact, so such a $\delta_j$ exists. We then also take $U_{\gamma_j}$ bounded and $\delta_j$ small enough that the closure of this set lies in the original $U_{\gamma_j}$ and $(s_j,\varpi_j)$ is injective on it, which compactness allows as well. On this set, $(s_j,\varpi_j)$ are local coordinates, transverse to the path and along it, as in \cite{akhtar2012path}.

\looseness=-1 The aerial manipulator must follow these paths and respect its physical limits. We impose joint limits $\underline\vartheta_i\le\vartheta_i\le\overline\vartheta_i$ with constants $\underline\vartheta_i<\overline\vartheta_i$, $i\in\{1,2,3\}$, and an attitude margin $|\phi|,|\theta|\le\pi/2-\epsilon$ with $\epsilon\in(0,\pi/2)$. We also impose thrust bounds $\utmin\le\ut\le\utmax$ with $0<\utmin<\utmax$, moment and joint torque bounds $|\tau_i|\le\overline\tau_i$ and $|\tau_{\vartheta,i}|\le\overline\tau_{\vartheta,i}$ with $\overline\tau_i,\overline\tau_{\vartheta,i}>0$, and a singularity margin $\detJ(\vartheta)\ge\epsilon_J$ with $\epsilon_J>0$.

\looseness=-1 Define the transverse errors $\alpha_c := s_c\circ\pc$, $\alpha_e := s_e\circ\pe$ and the path coordinates $\beta_c := \varpi_c\circ\pc$, $\beta_e := \varpi_e\circ\pe$, where $\pc$ and $\pe$ are read as functions of $x$ through $q$. Then $\alpha_c$ and $\alpha_e$ are smooth on $\Real^{20}$, and $\beta_c$ and $\beta_e$ are smooth on the open set $X_U := \{x\in\Real^{20} : \pc\in U_{\gamma_c},\ \pe\in U_{\gamma_e}\}$. The set $\{x : \alpha_c(x) = 0\}$ of states with the center of mass on $\gamma_c$ is not controlled invariant in general. As in \cite{akhtar2012path}, we aim to stabilize its largest controlled-invariant subset \cite[Sec.~3]{nielsen2008local}, the path-following manifold. Where the relative degrees of \cref{sec:normal} are well defined, it is $\mathcal{P}_c := \{x : L_F^k\alpha_c(x) = 0,\ k = 0,\dots,3\}$, and likewise $\mathcal{P}_e := \{x : L_F^k\alpha_e(x) = 0,\ k = 0,1\}$ for the end effector. We call $\mathcal{P} := \mathcal{P}_c\cap\mathcal{P}_e$ the dual path-following manifold. Along the paths we prescribe desired speeds $v_c$ and $v_e$ of the path coordinates and a desired heading $\psi_{\rm des}$, a constant or a smooth function of $\pc$. If $\psi_{\rm des}$ takes values in $\Real/2\pi\mathbb{Z}$, the heading error $\psi - \psi_{\rm des}$ is read in $(-\pi,\pi)$. Since the arm reaches only a bounded set of end-effector positions relative to the center of mass, the paths and the limits must satisfy the following assumption.

\begin{assumption}\label{ass:compat}
There exist $\ell_c^0\in D_c$, $\ell_e^0\in D_e$ and $\vartheta^0\in\Real^3$ with $\underline\vartheta_i<\vartheta_i^0<\overline\vartheta_i$ and $\detJ(\vartheta^0)>\epsilon_J$ such that $\sigma_e(\ell_e^0) - \sigma_c(\ell_c^0) = R_z\big(\psi_{\rm des}(\sigma_c(\ell_c^0))\big)\rho(\vartheta^0)$. Moreover, $\utmin<mg<\utmax$, and the input $(\tau,\tau_\vartheta) = -mg\,\Upsilon(\vartheta^0)$ lies strictly inside the moment and joint torque bounds.
\end{assumption}

\looseness=-1 The assumption asks for one admissible equilibrium on both paths. The equality says that some point of $\gamma_e$ is reachable from some point of $\gamma_c$ at level attitude with heading $\psi_{\rm des}$, at a nonsingular posture inside the joint limits. At rest there, the thrust $mg$ and the input $-mg\,\Upsilon(\vartheta^0)$ make both right-hand sides of \eqref{eq:model} vanish, so this equilibrium lies in $\mathcal{P}$ and satisfies every constraint strictly. We now state the dual path-following problem for an aerial manipulator.

\begin{problem}\label{prob:main}
Consider the aerial manipulator \eqref{eq:affine} with the paths $\gamma_c$ and $\gamma_e$, desired speeds $v_c$ and $v_e$, and desired heading $\psi_{\rm des}$, under \cref{ass:compat}. Design a state-feedback map $\bar u:\mathcal{U}\to\Real^7$, where $\mathcal{U}\subset\Real^{20}$ is open, and a set $\Omega_0\subset\mathcal{U}$ of admissible initial conditions such that, for every $x(0)\in\Omega_0$ and on the maximal interval of existence of its solution, the closed-loop system
\begin{enumerate}[leftmargin=*,nosep]
\item renders $\mathcal{P}$ invariant and locally exponentially stable,
\item makes the vehicle speed $\dot\beta_c$ and the heading $\psi$ converge exponentially to $v_c$ and $\psi_{\rm des}$,
\item satisfies $\underline\vartheta_i\le\vartheta_i(t)\le\overline\vartheta_i$, $|\tau_i(t)|\le\overline\tau_i$ and $|\tau_{\vartheta,i}(t)|\le\overline\tau_{\vartheta,i}$ for $i\in\{1,2,3\}$, $|\phi(t)|,|\theta(t)|\le\pi/2-\epsilon$, $\utmin\le x_{19}(t)\le\utmax$ and $\detJ(\vartheta(t))\ge\epsilon_J$.
\end{enumerate}
\end{problem}

\section{Normal Form of the Aerial Manipulator}\label{sec:normal}
\looseness=-1 We now choose the outputs to be linearized. The vehicle output is the center of mass $\pc$, not the body origin $p_b$. By \eqref{eq:com}, the acceleration of $\pc$ depends only on the thrust and the attitude, as for a quadrotor alone, so $\ud$ first appears in the fourth derivative of $\pc$. The torques already appear in the second derivative of $p_b = \pc - R\,r_c(\vartheta)$, through $r_c$, while $\ud$ appears in no second derivative. With $p_b$ as output, the column of $\ud$ in the decoupling matrix would be zero. We therefore define the virtual output map $\hat y: X_U\to\mathcal{Y}$, with $\mathcal{Y} := \Real^2\times\mathcal{C}_c\times\Real^2\times\mathcal{C}_e\times\Real$,
\begin{equation}\label{eq:vout}
\hat y(x) := \big(\alpha_c(x),\ \beta_c(x),\ \alpha_e(x),\ \beta_e(x),\ \psi\big),
\end{equation}
where $\psi = x_6$ is the yaw angle. Let $X_D := \{x\in X_U : x_{19}>0,\ |\phi|<\pi/2,\ |\theta|<\pi/2,\ \detJ(\vartheta)\ne0\}$. If $\psi_{\rm des}$ takes values in $\Real/2\pi\mathbb{Z}$, we also remove from $X_D$ the states with $\psi - \psi_{\rm des}(\pc) = \pi$ modulo $2\pi$, so that the heading error is smooth on $X_D$. The set $X_D$ is open. Let $D(x) := \big(L_GL_F^3\alpha_c,\ L_GL_F^3\beta_c,\ L_GL_F\alpha_e,\ L_GL_F\beta_e,\ L_GL_F\psi\big)\in\Real^{7\times7}$ be the decoupling matrix of $\hat y$.

\begin{proposition}\label[proposition]{prop:rd}
Consider the extended model \eqref{eq:affine} with the virtual output \eqref{eq:vout}. At every $x\in X_D$, the output has vector relative degree $\{4,4,4,2,2,2,2\}$ and
\begin{equation}\label{eq:detD}
\det D(x) = -\frac{x_{19}^2\cos\phi\ \det\Pi_c(\pc)\det\Pi_e(\pe)\ \detJ(\vartheta)}{m^3\cos\theta\ \det M_r(\vartheta)}.
\end{equation}
Moreover, with $\mathcal{Z} := \Real^8\times\mathcal{C}_c\times\Real^7\times\mathcal{C}_e\times\Real^3$, the map $T:X_D\to\mathcal{Z}$,
\begin{equation}\label{eq:T}
\begin{aligned}
T(x) := \big(&\alpha_c, L_F\alpha_c, L_F^2\alpha_c, L_F^3\alpha_c,\ \beta_c, L_F\beta_c, L_F^2\beta_c, L_F^3\beta_c,\\
&\alpha_e, L_F\alpha_e,\ \beta_e, L_F\beta_e,\ \psi, L_F\psi\big),
\end{aligned}
\end{equation}
is a local diffeomorphism at every point of $X_D$.
\end{proposition}
\begin{proof}
\looseness=-1 We differentiate each output until an input appears. By \eqref{eq:model}, no input appears in the first three derivatives of $\pc$ or in the first derivatives of $\pe = \pc + R\rho(\vartheta)$ and $\psi$, while
\begin{equation}\label{eq:snap}
\begin{aligned}
m\,\pc^{(4)} = {}& \ud Re_3 + 2x_{20}R\,\omega\times e_3\\
&+ x_{19}R\,\omega\times(\omega\times e_3) + x_{19}R\,\dot\omega\times e_3.
\end{aligned}
\end{equation}
Moreover, $\ddot\pe = \ddot\pc + R(\dot\omega\times\rho + J_\rho\ddot\vartheta) + R\,a_\rho(\vartheta,\omega,\dot\vartheta)$ for a smooth map $a_\rho$, and $\ddot\psi = Y_3\dot\omega + (\dot Y\omega)_3$, where $Y_3 := [0,\ \sin\phi/\cos\theta,\ \cos\phi/\cos\theta]$ is the third row of $Y$. By \eqref{eq:rot}, $(\dot\omega,\ddot\vartheta)$ is affine in $(\tau,\tau_\vartheta)$ with input matrix $M_r^{-1}$, so $u$ first appears in $\pc^{(4)}$, $\ddot\pe$ and $\ddot\psi$. Since $\alpha_c$, $\beta_c$, $\alpha_e$ and $\beta_e$ are $\pc$ and $\pe$ composed with the local coordinates $(s_c,\varpi_c)$ and $(s_e,\varpi_e)$, the chain rule gives the same orders for them, with the input terms multiplied by $\Pi_c$ and $\Pi_e$. It remains to show that $D$ is nonsingular on $X_D$.

In terms of $(\ud,\dot\omega,\ddot\vartheta)$, the input matrix of $(\pc^{(4)},\ddot\pe,\ddot\psi)$ is
\begin{equation*}
\hat D = \begin{bmatrix} R & 0 & 0\\ 0 & R & 0\\ 0 & 0 & 1\end{bmatrix}\begin{bmatrix} e_3/m & -(x_{19}/m)\hat e_3 & 0 \\ 0 & -\hat\rho & J_\rho \\ 0 & Y_3 & 0\end{bmatrix},
\end{equation*}
\looseness=-1 so $D = \mathrm{blkdiag}(\Pi_c,\Pi_e,1)\,\hat D\,\mathrm{blkdiag}(1,M_r^{-1})$. The first factor of $\hat D$ has unit determinant. In the second, the rows of $\pc^{(4)}$ are nonzero only in the columns of $\ud$, $\dot\omega_1$ and $\dot\omega_2$, so it is block lower triangular. The $3\times3$ block of these rows and columns has determinant $x_{19}^2/m^3$. The complementary $4\times4$ block, in the columns of $\dot\omega_3$ and $\ddot\vartheta$, has rows $[-\rho\times e_3\ \ J_\rho]$ and $(\cos\phi/\cos\theta)e_1^\top$, so its determinant is $-(\cos\phi/\cos\theta)\detJ$. Multiplying by $\det\Pi_c$, $\det\Pi_e$ and $\det M_r^{-1}$ gives \eqref{eq:detD}. Every factor in \eqref{eq:detD} is nonzero on $X_D$, so $D$ is nonsingular there and the vector relative degree is well defined. Finally, the relative degrees sum to $20 = \dim x$, so $T$ is a local diffeomorphism at every point of $X_D$ \cite[Sec.~5.1]{isidori1995nonlinear}.
\end{proof}

\looseness=-1 By \cref{prop:rd}, the extended model is locally fully feedback linearizable on $X_D$ and has no internal dynamics. The output replaces the joint angles in the flat outputs of \cite{welde2021dynamically, wei2021differential} by the end-effector path coordinates. Up to $\Pi_c$, $\Pi_e$ and $M_r^{-1}$, the matrix $D$ is therefore the decoupling matrix of their flat outputs times a block lower triangular factor with $1$, $I_3$ and $RJ_\rho$ on its diagonal. The thrust and attitude factors of \eqref{eq:detD} are thus those of the flat outputs, and the factor $\detJ$ is new and comes from the arm. We call a posture $\vartheta$ singular if $\detJ(\vartheta) = 0$, since $J_\rho$ then loses rank and the joints cannot move $\rho$ in every direction. To locate these postures, let $\varepsilon = \pm1$ according to whether the waist axis points along $\pm e_3$. Choose the first axis of $\mathcal{B}$ so that, at $\vartheta_1 = 0$, the upper arm and the forearm move in the plane spanned by $e_1$ and $e_3$. Every link beyond the waist rotates with it, so $\rho(\vartheta) = r_0 + R_z(\varepsilon\vartheta_1)\,\bar\rho(\vartheta_2,\vartheta_3)$ for a constant $r_0\in\Real^3$ and a smooth map $\bar\rho:\Real^2\to\Real^3$ with components $\bar\rho_x$, $\bar\rho_y$ and $\bar\rho_z$. Let $\Delta_\rho:\Real^2\to\Real$ be the determinant of $\partial(\bar\rho_x,\bar\rho_z)/\partial(\vartheta_2,\vartheta_3)$.

\begin{proposition}\label[proposition]{prop:arm}
For every $\vartheta\in\Real^3$, $\detJ(\vartheta) = -\varepsilon\,\bar\rho_x\,\Delta_\rho$, where $\Delta_\rho = c_\rho\sin(\vartheta_3 - \vartheta_3^*)$ for constants $c_\rho$ and $\vartheta_3^*$ set by the link geometry and masses.
\end{proposition}
\begin{proof}
\looseness=-1 The first column of $J_\rho$ is $\partial\rho/\partial\vartheta_1 = \varepsilon R_z(\varepsilon\vartheta_1)(e_3\times\bar\rho) = \varepsilon R_z(\varepsilon\vartheta_1)(-\bar\rho_y,\bar\rho_x,0)$, and the other two are $R_z(\varepsilon\vartheta_1)\,\partial\bar\rho/\partial\vartheta_j$, $j\in\{2,3\}$. The shoulder and the elbow rotate the upper arm and the forearm about axes normal to the plane spanned by $e_1$ and $e_3$, so these two columns have zero second component. Factoring out $R_z(\varepsilon\vartheta_1)$, which has unit determinant, and expanding along the second row, whose only nonzero entry is $\varepsilon\bar\rho_x$, gives the first claim. For the second claim, the upper arm rotates about the shoulder axis by $\vartheta_2$, and the forearm, with the bodies attached to it, rotates about the elbow axis by $\vartheta_2 + \vartheta_3$. Hence $\bar\rho = c_0 + R_y(\vartheta_2)c_2 + R_y(\vartheta_2+\vartheta_3)c_3$ for constants $c_0, c_2, c_3\in\Real^3$, with $c_2$ and $c_3$ in the plane spanned by $e_1$ and $e_3$. Thus $(\bar\rho_x,\bar\rho_z)$ is a planar two-link chain with link vectors $c_2$ and $c_3$, whose Jacobian determinant is $\pm\norm{c_2}\norm{c_3}\sin(\vartheta_3 - \vartheta_3^*)$, with $\vartheta_3^*$ the elbow angle at which $c_2$ and $R_y(\vartheta_3)c_3$ are parallel.
\end{proof}

\looseness=-1 When $\bar\rho_x = 0$, the waist moves $\rho$ only within the plane of the upper arm and the forearm, where the shoulder and the elbow already act. When $\vartheta_3 = \vartheta_3^*$ modulo $\pi$, the shoulder and the elbow move $\rho$ along a single direction. The barrier $\detJ\ge\epsilon_J$ of \cref{sec:barriers} keeps $\bar\rho_x$ and $\Delta_\rho$ from changing sign, so the end effector stays on one side of the plane $\bar\rho_x = 0$ with one elbow bend.

\looseness=-1 We now write the normal form. Define $\xi_j := L_F^{j-1}\alpha_{c,1}$, $\zeta_j := L_F^{j-1}\alpha_{c,2}$ and $\eta_j := L_F^{j-1}\beta_c$ for $j\in\{1,\dots,4\}$, and $\chi_j := L_F^{j-1}\alpha_{e,1}$, $\varrho_j := L_F^{j-1}\alpha_{e,2}$, $\varsigma_j := L_F^{j-1}\beta_e$ and $\mu_j := L_F^{j-1}\psi$ for $j\in\{1,2\}$, with $\alpha_{c,i}$, $\alpha_{e,i}$ the components of $\alpha_c$, $\alpha_e$. Thus $\xi,\zeta\in\Real^4$ and $\chi,\varrho\in\Real^2$ hold the transverse errors and their derivatives, $\eta\in\mathcal{C}_c\times\Real^3$ and $\varsigma\in\mathcal{C}_e\times\Real$ the path coordinates and their derivatives, and $\mu\in\Real^2$ the heading and its rate. These are the entries of $T(x)$ grouped by output, hence local coordinates on $X_D$ by \cref{prop:rd}. Let
\begin{equation}\label{eq:tfl}
u_{\rm TFL}(x) := D(x)^{-1}v(x),\qquad v := (v_1,\dots,v_7),
\end{equation}
where $(v_1,v_2) := -L_F^4\alpha_c + (v_\xi,v_\zeta)$, $v_3 := -L_F^4\beta_c + v_\eta$, $(v_4,v_5) := -L_F^2\alpha_e + (v_\chi,v_\varrho)$, $v_6 := -L_F^2\beta_e + v_\varsigma$ and $v_7 := -L_F^2\psi + v_\mu$, and $v_\xi,v_\zeta,v_\eta,v_\chi,v_\varrho,v_\varsigma,v_\mu\in\Real$ are auxiliary inputs. On $X_D$, in these coordinates, the feedback $u = u_{\rm TFL}(x)$ gives the normal form
\begin{equation}\label{eq:nf}
\begin{aligned}
&\dot\xi_j = \xi_{j+1},\ \dot\zeta_j = \zeta_{j+1},\ \dot\eta_j = \eta_{j+1},\ j\in\{1,2,3\},\\
&\dot\xi_4 = v_\xi,\ \dot\zeta_4 = v_\zeta,\ \dot\eta_4 = v_\eta,\ \dot\mu_1 = \mu_2,\ \dot\mu_2 = v_\mu,\\
&\dot\chi_1 = \chi_2,\ \dot\chi_2 = v_\chi,\ \dot\varrho_1 = \varrho_2,\ \dot\varrho_2 = v_\varrho,\ \dot\varsigma_1 = \varsigma_2,\ \dot\varsigma_2 = v_\varsigma.
\end{aligned}
\end{equation}

\looseness=-1 For constant speeds $v_c$ and $v_e$ and the desired heading $\psi_{\rm des}$, we choose the auxiliary inputs as
\begin{equation}\label{eq:vin}
\begin{aligned}
v_\xi &= -\textstyle\sum_{j=1}^4 k_{\xi j}\xi_j,\quad v_\zeta = -\textstyle\sum_{j=1}^4 k_{\zeta j}\zeta_j,\\
v_\chi &= -\textstyle\sum_{j=1}^2 k_{\chi j}\chi_j,\quad v_\varrho = -\textstyle\sum_{j=1}^2 k_{\varrho j}\varrho_j,\\
v_\eta &= -k_{\eta2}(\eta_2 - v_c) - k_{\eta3}\eta_3 - k_{\eta4}\eta_4,\ v_\varsigma = -k_{\varsigma}(\varsigma_2 - v_e),\\
v_\mu &= -k_{\mu1}(\mu_1 - \psi_{\rm des}) - k_{\mu2}(\mu_2 - \dot\psi_{\rm des}) + \ddot\psi_{\rm des},
\end{aligned}
\end{equation}
with gains for which the characteristic polynomials of the four transverse chains, as well as $s^3 + k_{\eta4}s^2 + k_{\eta3}s + k_{\eta2}$, $s + k_\varsigma$ and $s^2 + k_{\mu2}s + k_{\mu1}$, are Hurwitz. Here $\dot\psi_{\rm des} := L_F\psi_{\rm des}$ and $\ddot\psi_{\rm des} := L_F^2\psi_{\rm des}$ are the time derivatives of $\psi_{\rm des}$ under every input, since it depends on $x$ through $\pc$ and no input appears before $\pc^{(4)}$. Then $\mathcal{P}\cap X_D = \{x\in X_D : \xi = \zeta = \chi = \varrho = 0\}$ is invariant and locally exponentially stable, and the errors $\eta_2 - v_c$, $\varsigma_2 - v_e$ and $\mu_1 - \psi_{\rm des}$ decay exponentially.

\section{Proposed Controller}\label{sec:controller}
\looseness=-1 The controller \eqref{eq:tfl} is defined only on $X_D$ and ignores the joint and actuator limits. When the end-effector path demands a speed that the vehicle path cannot support, \eqref{eq:tfl} stretches the arm past a joint limit and into a singular posture, where $D$ is singular. A safety-filtered cascade, that is, \eqref{eq:tfl} followed by a CBF safety filter \cite{ames2019control}, prevents this. It may, however, change every row of $D(x)u = v(x)$, including the transverse rows, and then both points can leave their paths (\cref{sec:sim}). We instead keep six rows of $D(x)u = v(x)$ as equality constraints of a QP, namely the four transverse rows, the vehicle speed row and the heading row, and relax only the end-effector speed row. With a scalar slack $\delta\in\Real$ and $e_6\in\Real^7$ selecting the row of $\beta_e$, the controller is
\begin{subequations}\label{eq:qp}
\begin{align}
(u^\star,\delta^\star) = \argmin_{u\in\Real^7,\ \delta\in\Real}\ & \tfrac12\delta^2 \label{eq:qpa}\\
\text{s.t.}\quad & D(x)u = v(x) + e_6\delta, \label{eq:qpb}\\
& A(x)u \ge b(x), \label{eq:qpc}
\end{align}
\end{subequations}
where $A(x)\in\Real^{25\times7}$ and $b(x)\in\Real^{25}$ collect the ECBF rows and the actuator limits, which we define next.

\subsection{Barrier Functions and Actuator Limits}\label{sec:barriers}
\looseness=-1 We now define the rows of \eqref{eq:qpc}. Each limit in the third item of \cref{prob:main}, except the actuator bounds, becomes a barrier function $h$ that is nonnegative where the limit holds, and each barrier gives one ECBF row. For the joint limits, let $h_i^- := \vartheta_i - \underline\vartheta_i$ and $h_i^+ := \overline\vartheta_i - \vartheta_i$, $i\in\{1,2,3\}$. By \eqref{eq:rot}, $\ddot\vartheta$ depends on $u$ through the nonsingular $M_r^{-1}$, so these barriers have relative degree two. For the attitude margin, let $a_\epsilon := \pi/2-\epsilon$, $h_\phi^\pm := a_\epsilon\mp\phi$ and $h_\theta^\pm := a_\epsilon\mp\theta$. These have relative degree two by \eqref{eq:rot} and $\dot\Phi = Y(\Phi)\omega$. For the thrust bounds, let $h_t^- := x_{19} - \utmin$ and $h_t^+ := \utmax - x_{19}$, which have relative degree two with respect to $\ud$. The barrier $h_t^-$ also keeps the thrust positive, as $X_D$ requires. For the singularity margin, let $h_J := \detJ(\vartheta) - \epsilon_J$. Since $\ddot h_J = \nabla h_J^\top\ddot\vartheta + \dot\vartheta^\top\nabla^2h_J\,\dot\vartheta$, it has relative degree two wherever $\nabla h_J\ne0$. At a critical point of $\detJ$ its row has no input term and constrains the state alone, which the invariance statement of \cref{sec:ecbf} allows.

\looseness=-1 For each barrier $h$ with gains $(k_1,k_2)$, the row of \eqref{eq:qpc} is $L_GL_Fh\,u \ge -L_F^2h - k_1h - k_2L_Fh$, with $L_Gh = 0$ since $h$ depends only on $q$ or $x_{19}$. These are thirteen rows, and $\mathcal{A}_h$ denotes the admissible set of $h$. The actuator bounds are the remaining twelve rows, $\pm e_{i+1}^\top u\ge-\overline\tau_i$ and $\pm e_{i+4}^\top u\ge-\overline\tau_{\vartheta,i}$ for $i\in\{1,2,3\}$.

\subsection{Reduced Scalar QP and Regularity}\label{sec:reduced}
\looseness=-1 Eliminating $u$ through \eqref{eq:qpb} gives $u = D^{-1}(v + e_6\delta) = u_{\rm TFL} + \rdel\delta$, where $\rdel := D^{-1}e_6\in\Real^7$. The vector $\rdel$ is the input direction that leaves the six hard rows unchanged, since $D\rdel = e_6$. Substituting reduces \eqref{eq:qp} to
\begin{equation}\label{eq:rqp}
\delta^\star(x) = \argmin_{\delta\in\Real}\ \tfrac12\delta^2\quad\text{s.t.}\quad \bar a(x)\delta \ge \bar b(x),
\end{equation}
with $\bar a := A\rdel\in\Real^{25}$ and $\bar b := b - Au_{\rm TFL}\in\Real^{25}$, and the controller applies $u^\star(x) = u_{\rm TFL}(x) + \rdel(x)\delta^\star(x)$. The optimizer $\delta^\star$ is the feasible slack of least magnitude. It is zero whenever $u_{\rm TFL}$ satisfies \eqref{eq:qpc}, and then \eqref{eq:tfl} is applied.

Let $t_e := \Pi_e(\pe)^{-1}e_3\in\Real^3$ be the third column of $\Pi_e(\pe)^{-1}$, the direction in which $\varpi_e$ increases at unit rate while $s_e$ is stationary. We call the four attitude rows and the two thrust rows of \eqref{eq:qpc} the vehicle rows. The next lemma shows that the vehicle rows do not depend on the slack.
\begin{lemma}\label[lemma]{lem:zero}
For every $x\in X_D$, the six entries of $\bar a$ in \eqref{eq:rqp} that belong to the vehicle rows are zero, and the change of $(\ud,\dot\omega,\ddot\vartheta)$ per unit slack $\delta$ in \eqref{eq:qpb} is $(0,\,0,\,J_\rho^{-1}R^\top t_e)$.
\end{lemma}
\begin{proof}
\looseness=-1 By \eqref{eq:rot}, $(\ud,\dot\omega,\ddot\vartheta)$ is affine in $u$ with input matrix $\mathrm{blkdiag}(1,M_r^{-1})$, and $u = u_{\rm TFL} + \rdel\delta$, so its change per unit slack is $(\Delta\ud,\Delta\dot\omega,\Delta\ddot\vartheta) := \mathrm{blkdiag}(1,M_r^{-1})\,\rdel$. Since $D\rdel = e_6$ and $D = \mathrm{blkdiag}(\Pi_c,\Pi_e,1)\,\hat D\,\mathrm{blkdiag}(1,M_r^{-1})$ with $\Pi_c$ and $\Pi_e$ nonsingular on $X_D$, we have $\hat D(\Delta\ud,\Delta\dot\omega,\Delta\ddot\vartheta) = (0,\,t_e,\,0)$. The first three rows of $\hat D$ give $Re_3\,\Delta\ud + x_{19}R\,(\Delta\dot\omega\times e_3) = 0$. The two terms are orthogonal and $x_{19}\ne0$, so $\Delta\ud = 0$ and $\Delta\dot\omega_1 = \Delta\dot\omega_2 = 0$. The last row then gives $(\cos\phi/\cos\theta)\Delta\dot\omega_3 = 0$, and $|\phi|<\pi/2$ on $X_D$, so $\Delta\dot\omega = 0$. Rows $4$ to $6$ then give $RJ_\rho\Delta\ddot\vartheta = t_e$, and $\detJ\ne0$ on $X_D$, so $\Delta\ddot\vartheta = J_\rho^{-1}R^\top t_e$. Finally, the input terms of the attitude rows of \eqref{eq:qpc} are those of $\mp Y_j\dot\omega$, $j\in\{1,2\}$, with $Y_j$ the $j$th row of $Y$, and those of the thrust rows are $\mp\ud$. Their entries of $\bar a$ are therefore $\mp Y_j\Delta\dot\omega = 0$ and $\mp\Delta\ud = 0$.
\end{proof}

\looseness=-1 Since $\delta$ is scalar, the feasible set of \eqref{eq:rqp} is an interval. A row with $\bar a_i > 0$ bounds $\delta$ from below by $\bar b_i/\bar a_i$, a row with $\bar a_i < 0$ bounds it from above, and a row with $\bar a_i = 0$ holds if and only if $\bar b_i\le0$. Let $\underline\delta(x)$ be the largest lower bound and $\overline\delta(x)$ the smallest upper bound, or $\mp\infty$ when there is none. Then \eqref{eq:rqp} is feasible exactly when every row with $\bar a_i = 0$ holds and $\underline\delta\le\overline\delta$. Let $\Omega_S := \{x\in X_D : \bar a(x)\hat\delta > \bar b(x)\text{ for some }\hat\delta\in\Real\}$ be the set where \eqref{eq:rqp} is strictly feasible. It is open, since a strict inequality with fixed $\hat\delta$ persists near $x$ by continuity. Equivalently, $x\in\Omega_S$ exactly when every row with $\bar a_i = 0$ has $\bar b_i < 0$ and $\underline\delta < \overline\delta$. One pass through the $25$ rows gives the interval and both tests online.

\begin{theorem}\label{thm:lipschitz}
The QP \eqref{eq:rqp} has a unique minimizer at every $x\in\Omega_S$, given by $\delta^\star(x) = \max\{\underline\delta(x),\min\{0,\overline\delta(x)\}\}$, and $\delta^\star$ and $u^\star$ are locally Lipschitz functions of $x$ on $\Omega_S$.
\end{theorem}
\begin{proof}
\looseness=-1 The maps $D$, $v$, $A$ and $b$ are smooth on $X_D$, since $F$ and $G$ are smooth where $\cos\theta\ne0$, and $D$ is invertible there, so $\rdel$, $u_{\rm TFL}$, $\bar a$ and $\bar b$ are smooth on $X_D$. At $x\in\Omega_S$ the feasible set is a nonempty closed interval, and strong convexity gives the unique minimizer $\max\{\underline\delta,\min\{0,\overline\delta\}\}$, the point of the interval closest to zero. Fix $\bar x\in\Omega_S$. By continuity there is a neighborhood $U\subset\Omega_S$ of $\bar x$ on which each row with $\bar a_i(\bar x)\ne0$ keeps its sign and each row with $\bar a_i(\bar x) = 0$ has $\bar b_i < 0$. Let $\ell$ be the maximum of the smooth bounds $\bar b_i/\bar a_i$ over the rows with $\bar a_i(\bar x) > 0$ and $r$ the minimum over the rows with $\bar a_i(\bar x) < 0$, with an empty maximum $-\infty$ and an empty minimum $+\infty$. The point $\tilde\delta := \max\{\ell,\min\{0,r\}\}$ is continuous on $U$, hence bounded after shrinking $U$. On a smaller $U$, every row with $\bar a_i(\bar x) = 0$ holds strictly at $\tilde\delta$, since $\bar a_i$ tends to zero and $\bar b_i$ stays negative. Then $\tilde\delta$ is feasible and minimizes $|\delta|$ over an interval that contains the feasible set, so $\delta^\star = \tilde\delta$ on $U$. This is locally Lipschitz, and so is $u^\star = u_{\rm TFL} + \rdel\delta^\star$.
\end{proof}
The closed-loop vector field is therefore locally Lipschitz on $\Omega_S$, and from every $x(0)\in\Omega_S$ there is a unique solution in $\Omega_S$ on a maximal interval $[0,t_f)$.

\subsection{Path Invariance and Safety}
\looseness=-1 We now show that the constraints act only on the end-effector speed chain $\varsigma$. Let $z_\perp := (\xi,\zeta,\chi,\varrho)\in\Real^{12}$ be the transverse coordinates, let $z_v := (\eta_2 - v_c,\ \eta_3,\ \eta_4,\ \mu_1 - \psi_{\rm des},\ \mu_2 - \dot\psi_{\rm des})\in\Real^5$ be the vehicle speed and heading errors, and let $z := (z_\perp, z_v)\in\Real^{17}$. Let $\Lambda := \mathrm{blkdiag}(\Lambda_\xi,\Lambda_\zeta,\Lambda_\chi,\Lambda_\varrho,\Lambda_\eta,\Lambda_\mu)$, where each block is the companion matrix of the Hurwitz polynomial of its chain in \eqref{eq:vin}. Then $\Lambda$ is Hurwitz.
\begin{lemma}\label[lemma]{lem:invariance}
Along every closed-loop solution in $\Omega_S$, $\dot z = \Lambda z$.
\end{lemma}
\begin{proof}
By \cref{prop:rd}, under any input $u$ each chain of \eqref{eq:nf} is a chain of integrators whose last state has derivative $L_F^ry + D_iu$, with $y$ its output, $r$ its relative degree and $D_i$ its row of $D$. By \eqref{eq:qpb}, $Du^\star = v + e_6\delta^\star$, so each last derivative equals its auxiliary input, except that $\dot\varsigma_2 = v_\varsigma + \delta^\star$. Since $v_c$ is constant and $\dot\psi_{\rm des}$ and $\ddot\psi_{\rm des}$ are the derivatives of $\psi_{\rm des}$, \eqref{eq:vin} gives $\dot z = \Lambda z$.
\end{proof}
The solution exists only while it stays in $\Omega_S$, where \eqref{eq:rqp} is strictly feasible. By \cref{lem:zero}, each vehicle row reads $\bar b_i(x)\le0$, so the slack cannot enforce it. We say that the row holds strictly at $x$ if $\bar b_i(x) < 0$, as $\Omega_S$ requires. The next lemma shows that the vehicle rows hold strictly near the nominal vehicle motion at low vehicle speed.
\begin{lemma}\label[lemma]{lem:vehicle}
Let $\gamma_c$ be closed. Under \cref{ass:compat}, there exist $\epsilon_z,\bar v_z>0$ such that the vehicle rows hold strictly at every $x\in X_D$ with $\norm{z(x)} < \epsilon_z$ whenever $|v_c|<\bar v_z$.
\end{lemma}
\begin{proof}
\looseness=-1 For a vehicle row with barrier $h$, $\bar b_i = -(L_F^2h + L_GL_Fh\,u_{\rm TFL}) - k_1h - k_2L_Fh$. It depends on $\Phi$, $\omega$, $x_{19}$, $x_{20}$ and the values of $\ud$ and $\dot\omega$ under $u_{\rm TFL}$. By \eqref{eq:nf}, this feedback gives $\alpha_c^{(4)} = (v_\xi,v_\zeta)$, $\beta_c^{(4)} = v_\eta$ and $\ddot\psi = v_\mu$. Through $(s_c,\varpi_c)$, the point $\pc$ and its first four derivatives are therefore functions of $(\beta_c,\xi,\zeta,z_v,v_c)$, and so are $\psi = \psi_{\rm des} + z_{v,4}$, $\dot\psi = \dot\psi_{\rm des} + z_{v,5}$ and $\ddot\psi$. On $X_D$ the third entry of $Re_3$ is $\cos\phi\cos\theta>0$. Hence \eqref{eq:com} and its derivative give $x_{19}$, $x_{20}$, $\Phi$ and $\omega$ as smooth functions of $\ddot\pc$, $\dddot\pc$, $\psi$ and $\dot\psi$ wherever the third entry of $\ddot\pc + ge_3$ is positive. Since $x_{19}\ne0$ and $\cos\phi\ne0$, \eqref{eq:snap} and $\ddot\psi = Y_3\dot\omega + (\dot Y\omega)_3$ then give $\ud$ and $\dot\omega$. Thus, on $X_D$, $\bar b_i = g_i(\beta_c,\xi,\zeta,z_v,v_c)$ for a continuous map $g_i$ on an open set $W\subset\mathcal{C}_c\times\Real^{14}$. At $(\xi,\zeta,z_v,v_c) = 0$ the center of mass rests on $\gamma_c$, so $x_{19} = mg$, $\Phi = (0,0,\psi_{\rm des})$, $\omega = 0$, $x_{20} = 0$, $\ud = 0$ and $\dot\omega = 0$. There, $\bar b_i = -k_1h<0$, since $a_\epsilon>0$ and $\utmin<mg<\utmax$ by \cref{ass:compat}. The compact set $\mathcal{C}_c\times\{0\}$ lies in $W$, since $\ddot\pc + ge_3 = ge_3$ there. By the tube lemma, there is $\epsilon_z > 0$ such that every $g_i$ is negative wherever $\norm{(\xi,\zeta,z_v)} < \epsilon_z$ and $|v_c| < \epsilon_z$. Set $\bar v_z := \epsilon_z$.
\end{proof}
\looseness=-1 Let $S\succ0$ solve $\Lambda^\top S + S\Lambda = -I$, let $\underline s>0$ be its smallest eigenvalue, and let $V := z^\top S z$. Fix $\kappa>0$ with $\kappa\le\underline s\min\{\delta_c,\delta_e,\pi\}^2/4$ and, when $\gamma_c$ is closed, $\kappa<\underline s\epsilon_z^2$. Let $\Omega_\kappa := \{x\in X_D : V(x) < \kappa\}$ and $\bar\Omega := X_D\cap\bigcap_h\mathcal{A}_h$. Since $\alpha_c$, $\alpha_e$ and $\mu_1 - \psi_{\rm des}$ are entries of $z$ and $\norm{z}\le(V/\underline s)^{1/2}$, every $x\in\Omega_\kappa$ has $\norm{s_c(\pc)} < \delta_c/2$, $\norm{s_e(\pe)} < \delta_e/2$, $|\mu_1 - \psi_{\rm des}| < \pi/2$ and, when $\gamma_c$ is closed, $\norm{z}<\epsilon_z$. By \cref{lem:vehicle}, when $\gamma_c$ is closed and $|v_c|<\bar v_z$, such an $x$ lies in $\Omega_S$ if and only if the other $19$ rows are strictly feasible at it. These rows can fail inside $\Omega_\kappa\cap\bar\Omega$, since $z$ contains neither the position of the end effector along its path nor its speed. Their strict feasibility is therefore a property of the solution, not of the initial condition. Feasibility along the solution is the core assumption of CBF-based controllers, and checking it for several barriers under input bounds is difficult in general \cite[Sec.~I]{tan2022compatibility}. We assume it for these $19$ rows with a uniform margin. For $w,\delta_M>0$, let $\Omega_S^w$ be the set of $x\in X_D$ at which some $\hat\delta$ with $|\hat\delta|\le\delta_M$ satisfies $\bar a_i(x)\hat\delta\ge\bar b_i(x) + w$ for each of these rows. It is closed relative to $X_D$, since the bounded $\hat\delta$ have convergent subsequences and the inequalities pass to the limit.
\begin{assumption}\label{ass:slater}
There exist $w,\delta_M>0$ such that the closed-loop solution stays in $\Omega_S^w$ on $[0,t_f)$.
\end{assumption}

\looseness=-1 Let $\Omega_0 := \Omega_\kappa\cap\bar\Omega\cap\Omega_S$ be the set of admissible initial conditions of \cref{prob:main}. To show that it is nonempty, let $\psi^0 := \psi_{\rm des}(\sigma_c(\ell_c^0))$, let $x^0\in\Real^{20}$ be the state with $p_b = \sigma_c(\ell_c^0) - R_z(\psi^0)r_c(\vartheta^0)$, $\Phi = (0,0,\psi^0)$, $\vartheta = \vartheta^0$, $\dot q = 0$, $x_{19} = mg$ and $x_{20} = 0$, and let $u^0 := (0,\,-mg\,\Upsilon(\vartheta^0))$.
\begin{lemma}\label[lemma]{lem:omega0}
\looseness=-1 Under \cref{ass:compat}, there exists $\bar v>0$ such that $\Omega_0$ contains a neighborhood of $x^0$ whenever $|v_c|<\bar v$, for every $v_e\in\Real$.
\end{lemma}
\begin{proof}
\looseness=-1 Since $\pc = p_b + Rr_c$, we have $\pc = \sigma_c(\ell_c^0)\in\gamma_c$, and \cref{ass:compat} gives $\pe = \pc + R_z(\psi^0)\rho(\vartheta^0) = \sigma_e(\ell_e^0)\in\gamma_e$. With $x_{19} = mg>0$, $\phi = \theta = 0$, $\psi = \psi_{\rm des}(\pc)$ and $\detJ(\vartheta^0)>\epsilon_J$, this gives $x^0\in X_D$. Since $Re_3 = e_3$ and $c$ vanishes at zero velocity, both right-hand sides of \eqref{eq:model} vanish under $u^0$, so $x^0$ is an equilibrium. At an equilibrium every output derivative vanishes, so $Du^0 = -\big(L_F^4\alpha_c, L_F^4\beta_c, L_F^2\alpha_e, L_F^2\beta_e, L_F^2\psi\big)(x^0)$, $\dot\psi_{\rm des} = \ddot\psi_{\rm des} = 0$ and $z(x^0) = -v_ce_{13}$. By \eqref{eq:vin}, the only nonzero auxiliary inputs at $x^0$ are $v_\eta = k_{\eta2}v_c$ and $v_\varsigma = k_\varsigma v_e$. With $\hat\delta := -k_\varsigma v_e$, \eqref{eq:tfl} then gives $u_{\rm TFL}(x^0) + \rdel\hat\delta = u^0 + k_{\eta2}v_cD(x^0)^{-1}e_3$. Each barrier $h$ has $h(x^0)>0$ by \cref{ass:compat} and $\epsilon<\pi/2$, and its first two derivatives vanish at the equilibrium. Its row of \eqref{eq:qpc} at $u^0$ therefore reads $0\ge-k_1h(x^0)$ and holds strictly, and $x^0$ is interior to $\mathcal{A}_h$ since $\nu_1<0$. The actuator rows hold strictly at $u^0$ by \cref{ass:compat}. Hence $x^0\in\bar\Omega$. Since $\bar a\hat\delta - \bar b = A(u_{\rm TFL} + \rdel\hat\delta) - b$, the slack $\hat\delta$ satisfies every row of \eqref{eq:rqp} strictly when $v_c = 0$. These finitely many strict inequalities are continuous in $v_c$ and do not involve $v_e$, and $V(x^0) = S_{13,13}v_c^2$, so some $\bar v>0$ gives $x^0\in\Omega_S\cap\Omega_\kappa$ whenever $|v_c|<\bar v$. Since $\Omega_S$ and $\Omega_\kappa$ are open and $x^0$ is interior to $\bar\Omega$, $\Omega_0$ contains a neighborhood of $x^0$.
\end{proof}
We can now state the main result of the paper.

\begin{theorem}\label{thm:safety}
Under \cref{ass:compat}, every closed-loop solution under \eqref{eq:qp} with $x(0)\in\Omega_0$ satisfies the third item of \cref{prob:main} on $[0,t_f)$. Moreover, there exist $C_1,C_2>0$, independent of $x(0)$, such that $\norm{z_\perp(t)}\le C_1e^{-C_2t}\norm{z_\perp(0)}$ and $\norm{z_v(t)}\le C_1e^{-C_2t}\norm{z_v(0)}$ for all $t\in[0,t_f)$. In particular, $x(0)\in\mathcal{P}$ implies $x(t)\in\mathcal{P}$ for all $t\in[0,t_f)$. If in addition $\gamma_c$ and $\gamma_e$ are closed, $|v_c|<\bar v_z$ and \cref{ass:slater} holds, then $t_f = \infty$.
\end{theorem}
\begin{proof}
\looseness=-1 On $[0,t_f)$ the solution lies in $\Omega_S$, so every row of \eqref{eq:qpc} holds along it. Since $x(0)\in\bar\Omega$, the invariance statement of \cref{sec:ecbf} gives $x(t)\in\mathcal{A}_h$ for every barrier $h$. This yields the third item of \cref{prob:main}, with the actuator rows giving the moment and joint torque bounds. By \cref{lem:invariance}, $\dot V = -\norm{z}^2$ along the solution, so $V(x(t))\le V(x(0))<\kappa$ and $x(t)\in\Omega_\kappa$. Since $\Lambda$ is Hurwitz and block diagonal, $z_\perp$ and $z_v$ satisfy the stated bounds with any $C_1,C_2>0$ such that $\norm{e^{\Lambda t}}\le C_1e^{-C_2t}$. With $z_\perp(0) = 0$, the first bound gives $z_\perp(t) = 0$, which is the invariance of $\mathcal{P}$ since $\mathcal{P}\cap X_D = \{x\in X_D : z_\perp(x) = 0\}$.

\looseness=-1 Now let $\gamma_c$ and $\gamma_e$ be closed, let $|v_c|<\bar v_z$ and \cref{ass:slater} hold, and suppose $t_f<\infty$. The barriers bound $\vartheta$, $\phi$, $\theta$ and $x_{19}$, and the second inequality of each admissible set bounds $\dot\vartheta$, $\dot\phi$, $\dot\theta$ and $x_{20}$. Since $x(t)\in\Omega_\kappa$, the points $\pc$ and $\pe$ stay in the sets $K_j := \{p\in U_{\gamma_j} : \norm{s_j(p)}\le\delta_j/2\}$, $j\in\{c,e\}$, which are compact because the paths are closed and the closure of $U_{\gamma_j}$ lies in a tubular neighborhood of $\gamma_j$. On $K_c$ the matrix $\Pi_c^{-1}$ and the gradient of $\psi_{\rm des}$ are bounded, and so are $\xi_2$, $\zeta_2$ and $\eta_2$ by the bound on $z$. Hence $\dot\pc$ and $\dot\psi = \mu_2$ are bounded. In turn, $\dot\Phi$ is bounded, and so is $\omega = Y(\Phi)^{-1}\dot\Phi$, since $|\theta|\le\pi/2-\epsilon$. Then $p_b = \pc - Rr_c$ and $\dot p_b$ are bounded. Thus $x(t)$ is bounded on $[0,t_f)$, with $\psi$ bounded because $t_f<\infty$, and it has a limit point $\bar x$ as $t\to t_f$. The closed conditions $x_{19}\ge\utmin$, $|\phi|,|\theta|\le\pi/2-\epsilon$, $\detJ\ge\epsilon_J$ and $|\mu_1 - \psi_{\rm des}|\le\pi/2$ and the compact sets $K_c$ and $K_e$ give $\bar x\in X_D$. By \cref{ass:slater} and since $\Omega_S^w$ is closed relative to $X_D$, also $\bar x\in\Omega_S^w$. Since $V(\bar x)\le\kappa<\underline s\epsilon_z^2$, \cref{lem:vehicle} gives the vehicle rows strictly at $\bar x$, so $\bar x\in\Omega_S$. This contradicts the maximality of $t_f<\infty$, since the solution must eventually leave every compact subset of the open set $\Omega_S$. Hence $t_f = \infty$.
\end{proof}
\looseness=-1 By \cref{lem:omega0}, $\Omega_0$ contains a neighborhood of $x^0$ when $|v_c|<\bar v$. The controller \eqref{eq:qp} on $\mathcal{U} := \Omega_S$ with $\Omega_0$ then solves \cref{prob:main}. \cref{thm:safety} gives the third item of \cref{prob:main} directly, and its bounds on $z_\perp$ and $z_v$ give the first two items, since $\mathcal{P}\cap X_D = \{z_\perp = 0\}$, $\dot\beta_c = \eta_2$ and $\psi = \mu_1$. If \eqref{eq:rqp} becomes infeasible, the interval test detects it. A fallback with safety guarantees is left to future work.

\section{Simulation Results}\label{sec:sim}
\looseness=-1 We simulate in Drake \cite{drake} a Skydio X2 quadrotor from the MuJoCo Menagerie \cite{menagerie2022} carrying an OpenMANIPULATOR-X arm below the airframe (\cref{fig:concept}). The first three joints are actuated, the wrist is fixed, and the combined system weighs $1.917$\,kg. The upper arm is bent, so $\detJ$ vanishes at $\vartheta_3 = \vartheta_3^* = -1.38$\,rad, and the set $\bar\rho_x = 0$ crosses the joint box. The joint limits stop $0.08$\,rad short of $\vartheta_3^*$ but do not exclude $\bar\rho_x = 0$, so the barrier $h_J$ is needed. The controller uses the exact multibody model at $400$\,Hz with a zero-order hold, and the plant is integrated at $4$\,kHz, so the runs are sampled approximations of \eqref{eq:qp}. The \href{https://gradslab.github.io/am_dual_pf-/}{project page} provides the code, the animations, all settings, and studies of the sampling rate, the compatible postures and model mismatch.

\begin{figure}[t]
\centering
\includegraphics[width=\linewidth]{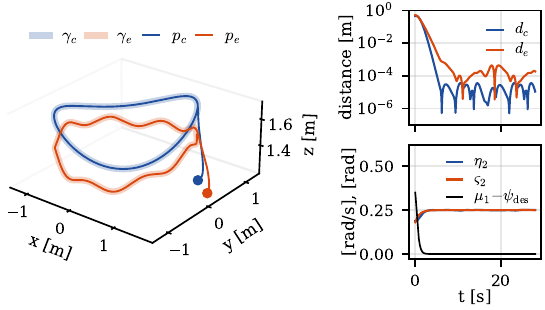}

\vspace{-0.15in}

\caption{Nominal controller from an off-path start (dots). Left: center of mass (blue), end effector (red) and paths (wide light lines). Right: distances to the paths, along-path speeds and heading error.}
\label{fig:nominal}

\vspace{-0.14in}

\end{figure}

\looseness=-1 The vehicle path is a horizontal circle of radius $1.2$\,m with a wavy altitude, $s_c = (x^2+y^2-1.2^2,\ z - 1.6 - 0.1\sin 2\lambda)$ with $\lambda := \mathrm{atan2}(y,x)$ and $\varpi_c = \lambda$. The end-effector path is the circle of the same radius below it with a scalloped altitude, $s_e = (x^2+y^2-1.2^2,\ z - 1.42 - 0.06\sin(2\lambda+0.8) - 0.03\sin 8\lambda)$ and $\varpi_e = \lambda$. Both maps are smooth away from the $z$ axis, which $U_{\gamma_c}$ and $U_{\gamma_e}$ avoid. The heading reference is the tangent direction $\psi_{\rm des} = \lambda + \pi/2$, and a speed of $0.25$\,rad/s is $0.3$\,m/s. We report $d_c$ and $d_e$, the distances of $\pc$ and $\pe$ to the points of their paths at the same $\lambda$, which bound the distances to the paths. The limits are $\epsilon = 0.1$\,rad, $\utmin = 2$\,N, $\utmax = 40$\,N, $\overline\tau = (3,3,1)$\,Nm, $\overline\tau_{\vartheta,i} = 3$\,Nm, $\epsilon_J = 1.5\times10^{-3}$\,m$^3$ and joint limits $\pm1.5$, $\pm1.3$ and $[-1.3,1.2]$\,rad. The simulated plant does not enforce the joint limits, so only the barriers do. All runs start $0.4$\,m outside and $0.25$\,m below the vehicle path, moving at $0.3$\,m/s with a heading error of $0.35$\,rad, inside every $\mathcal{A}_h$ but off $\mathcal{P}$. We do not compute $\kappa$, so the runs test the controller beyond the certified set.

\begin{figure}[t]
\centering
\includegraphics[width=\linewidth]{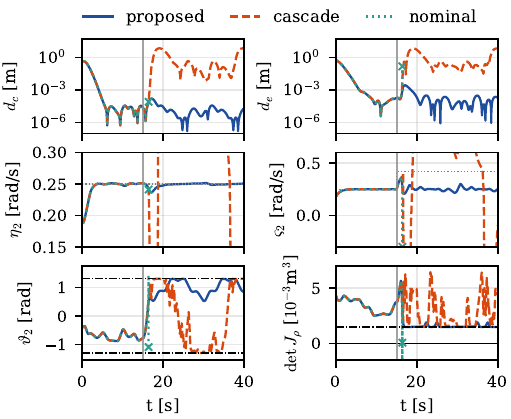}

\vspace{-0.15in}

\caption{Incompatible speed references: distances to the paths, along-path speeds with references (gray dotted), $\vartheta_2$ and $\detJ$ with limits and margin (black dash-dotted). The gray line marks the switch at $t = 15$\,s. The nominal run diverges at the cross, and the cascade speeds leave the plotted range.}
\label{fig:relaxed}

\vspace{-0.08in}

\end{figure}

\looseness=-1 Under the nominal controller \eqref{eq:tfl} with $v_c = v_e = 0.25$\,rad/s (\cref{fig:nominal}), $d_c$ and $d_e$ decay exponentially from $0.47$\,m and $0.53$\,m and fall below $1$\,mm after $5.8$\,s. They then stay below $0.04$\,mm and $1$\,mm, the sampled-data residual of the invariance of $\mathcal{P}$. The speeds settle at $0.25$\,rad/s and the heading error below $10^{-4}$\,rad. \cref{fig:relaxed} compares \eqref{eq:qp} with the nominal controller and the safety-filtered cascade, which applies the filter of \cite[Sec.~II-C]{ames2019control} with the rows of \eqref{eq:qpc} to \eqref{eq:tfl}. The cascade minimizes $\tfrac12\norm{u - u_{\rm TFL}}^2$, an identity weight on inputs of mixed units, subject to these rows. At $t = 15$\,s the end-effector reference is raised to $v_e = 0.42$\,rad/s with $v_c$ unchanged. The three controllers coincide until the switch, after which the nominal controller stretches the arm until $\detJ$ crosses zero at $t = 16.39$\,s.

\looseness=-1 With \eqref{eq:qp}, the singularity row activates at $t = 16.26$\,s, and the slack brings the end-effector speed back to the vehicle speed with $|\delta^\star|\le1.04$\,rad/s$^2$. From then on the singularity row or the upper limit of $\vartheta_2$ is active at every step, so the closed loop moves along the constraint boundary. The barrier holds up to a residual of $3\times10^{-6}$\,m$^3$, and after $t = 20$\,s, $\detJ$ stays below $1.93\times10^{-3}$\,m$^3$, close to $\epsilon_J$. The distances peak at $2.8$\,mm for $d_e$ and $0.16$\,mm for $d_c$, and after $t = 20$\,s they are back below $0.32$\,mm and $0.04$\,mm.

\looseness=-1 After $t = 20$\,s the end-effector speed has mean $0.250$\,rad/s and range $[0.216, 0.300]$\,rad/s. The vehicle speed stays in $[0.236, 0.252]$\,rad/s, a sampled-data dip, since the hard rows hold it at $0.25$\,rad/s in continuous time. The heading error stays below $10^{-4}$\,rad, and the thrust, moments and joint torques stay well inside their bounds. The reduced QP stays strictly feasible, with an interval at least $2.6$\,rad/s$^2$ wide, more than twice the largest slack, and the vehicle rows hold with large margins, as \cref{lem:vehicle} predicts. Under the cascade, the end effector leaves its path by over $5$\,cm within $0.25$\,s of its first active row, and the vehicle after $0.7$\,s. The vehicle speed reverses to $-4.2$\,rad/s with the moment bounds and $\utmax$ active, and neither point returns within $7$\,mm of its path.

\section{Conclusion}
\looseness=-1 This paper presented a safe dual path-following controller for a quadrotor carrying a 3-DOF arm. With the center of mass of the combined system as the vehicle output, the dynamically extended system is feedback linearizable away from an explicit singular set. A QP that relaxes only the end-effector speed row reduces to a scalar program with a closed-form solution. While it stays strictly feasible, the feedback is locally Lipschitz, both paths are invariant and every constraint holds. In Drake, the controller resolved incompatible speed references by slowing the end effector, while the cascade with a safety filter left both paths and the nominal controller reached the singular set. Future work includes collision avoidance, a verifiable feasibility condition, a guaranteed fallback and hardware tests.

\textit{Generative AI Disclosure: } ChatGPT 5.6 has been used to improve the syntax and grammar of some paragraphs.

\bibliographystyle{IEEEtran}
\bibliography{references}

\end{document}